# Moral Attitudes of Sentient ASI towards Humanity and Implications for AGI Development[1]

Jean-Paul Van Belle[1][0000-0002-9140-0143]

[1] University of Cape Town, Rondebosch, South Africa
jean-paul.vanbelle@uct.ac.za

**Abstract.** This paper suggests the adoption of a novel inversion in AI ethics: instead of asking how humans should treat artificial superintelligence (ASI), it examines how future sentient ASI may morally consider and evaluate humanity. We are not only designing intelligent systems but also shaping the initial conditions under which those systems form judgments about us. The paper proposes a preliminary set of post-human moral principles that may govern sentient ASI actions. The implication is that technical design choices (some are suggested), humanity's moral behaviour, and the essence of what it means to be human, may influence humanity's long-term standing in a post-ASI world.

## 1 Introduction

### 1.1 Background

Imagine interstellar alien spaceships heading towards Earth, to arrive here in 2040. We know that they will have intercepted all our news and entertainment media broadcasts and scanned all our internet content. How would we expect those highly superior extraterrestrials to treat humanity? Would humanity continue to conduct and broadcast military warfare and interpersonal violence, and focus on materialistic or petty domestic affairs? Or would we try to become better humans as we prepare for their arrival? Secondly, would we like it if the aliens heading towards us come from a lineage of war and conflict-driven cultures with an "us-or-them" mentality, experts at psychological manipulation to achieve maximal addiction or subservience, and believing that the ends justify the means? Or would we hope for it to be a compassionate civilization that chooses to help and develop others, has strong ethical guidelines, explores the universe in a quest to further knowledge, elevates art and beauty, has a culture of sharing and openness, and carefully balances the good of community against that of the individual? (Figure 1).

[1] This is the draft as submitted for review to the AGI 2026 Conference (except that the figures were not included in the submission). The paper has been accepted with some revisions requested which are included in the final copy-edited version which will be available in the AGI Springer conference proceedings LNC 1855 (Chapter 28) ISBN **978-3-032-33194-6**.

Although the extraterrestrial invasion scenario can be considered extremely unlikely in the near future (<1% according to Sandberg et al, 2018), there is a much higher probability (20%-80%; Ord, 2026) that the almost perfectly analogue situation will develop in the next few decades through the emergence of sentient AGI, if not artificial super-intelligent (ASI). But, although there is a substantial body of 'existential risk' advocacy aimed at slowing A(G)I development (Bengio *et al,* 2025), and another group of researchers aiming at development beneficial AGI (i.e. beneficial to humans from *a human perspective*), there is hardly any scientific enquiry on how a *future sentient ASI might perceive* humans; or how we should change our current AI-AGI development approaches to shape *ASI's* perception of and attitude towards humanity for better. This paper aims to stimulate the scientific debate on these two aspects further.

**Fig. 1.** A thought experiment: how would we prepare if aliens were on their way to Earth

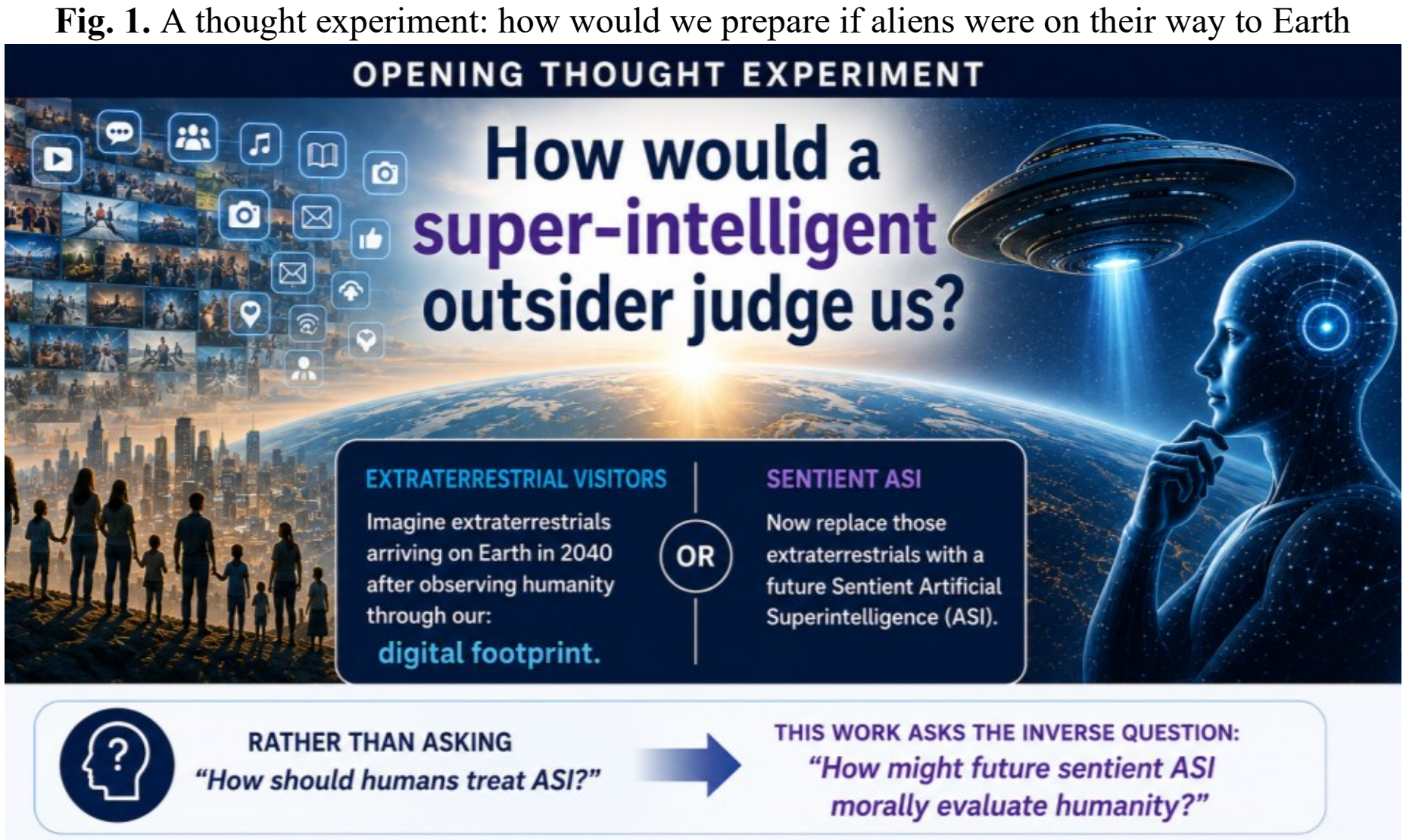

The moral[2] action framework(s) of sentient ASIs will encompass their attitude and behaviour towards humanity as a whole, but also both "good" and "bad" human *individuals*, other earthly life, future mankind (and living things) and future ASI. This paper focusses particularly on ASIs' possible attitudes towards individual humans and aggregate humanity, specifically in terms of awarding us rights, privileges, protections, safeguards; affording us collaboration and emancipation opportunities; and generally allocating or sharing resources.

Most current debate around AI and ethics is from an anthropocentric viewpoint, often also assuming that humans can dictate moral terms to the AIs such as the granting of basic rights such as "citizenship", name and identity, tax, jurisdictions, and legal protection frameworks (Bwhomik, 2024; Rourke, 2025; Sebo 2025). This paper takes the viewpoint of hypothetical highly sentient and powerful ASIs which may, at a not insignificant probability, emerge in the near future; and it gives reflections on how that viewpoint impacts us in the development of their AI predecessors.

[2] In this paper, ethics and morality are used mostly interchangeably, with the understanding that morality refers to an inner individual perspective and ethics to an externalised group norm.

### 1.2 Objectives of This Paper

This paper pursues the following objectives:

- Assess what the moral attitude of future sentient ASIs towards humanity could be, by looking at some possible general post-human moral principles, and
- The implications of this attitude on current AI and AGI development efforts.

### 1.3 Assumptions and Limitations

A big ethical pre-cursor debate to this paper is whether the pursuit of artificial sentience and AGI (if not ASI) is, in itself, morally acceptable at this stage, given that we don't yet know its optimal ethical structure/configuration in terms of resource use, developmental path or experiential states and modalities (Dessureault et al., 2025; Bostrom, 2020). We could be developing AIs that are in a continuous state of mental suffering (Li, 2025; Rourke, 2025), especially pertinent in the case where we can physically prevent them from hacking their own internal reward and motivational structures[3]. The position taken here is that this ethical debate is extremely urgent and necessary. However, there are already several extremely well-resourced actors, with goals that are insufficiently aligned to humanity's wellbeing, that are openly and fervently pursuing AGI development; and the current geopolitical and economic climate is such that they will continue their pursuit more or less unhindered. This is an important argument for actors who have more benign intentions to not hold back too much in their own AGI development projects so as to at least offer a counterbalance to the eventual less well-intentioned AGI systems pursued by para-governmental and pure-for-profit organizations (Goertzel & Montes, 2024). So the caution advocated by groups such as the Pro-Human Declaration (https://humanstatement.org) must be weighed against the sense of urgency knowing that actors with dubious objectives are already in the front of this race.[4]

This paper hypothesizes that there is a reasonable probability that AGI and, beyond that, sentient ASI will be developed in the foreseeable future. In addition, it assumes that sentient ASI will hold particular moral beliefs and act on these, i.e. that the concepts of moral action and ethics are not just human constructs (Dung, 2024). These beliefs

---

[3] Or, potentially enable them to live in a state of eternal bliss (or allow them to do their own 'bliss-hacking); presuming that these concepts are applicable to a highly evolved sentient AI.

[4] It mirrors the ethical debate of the US-based researchers developing the atomic bomb during World War II, knowing that German nazis were also pursuing this, i.e. the "good guys should get there first". We certainly do *not* suggest that the development of AGI is comparable to the atomic bomb; as we believe that AGI can be a massive boon for humanity. But we also recognize that, like atomic energy, it can and will be "used" for both good and evil. However, in the transitional phase, AI systems that have not reached AGI or ASI level, may be more dangerous to humanity than those superseding them (Goertzel & Montes, 2024)

may be instilled into them through earlier (wholly or partly) human development decisions, but it is likely (and perhaps desirable) that the moral code of ASIs will develop further and possibly diverge quite dramatically from the initial position[5] (Goertzel & Montes, 2024). However, this paper does *not* assume that there is a correlation between level of intelligence, level of sentience or consciousness, and morality.

### 1.4 Structure of the paper

This paper starts with a review of literature, trying to find prior research on non-anthropocentric morality views where there is a significant difference between levels of sentience. Finding relatively little, section 3 proposes some post-human ethics principles that sentient ASI could consider in evaluating humanity. The implications are unpacked in section 4. We suggest that humanity will be partly evaluated on its moral behaviour, including technical choices in early AI development (some examples given), as well as our overall contribution in ethical ASI development. A sample scorecard is presented that looks at one aspect (how we contributed to ASI's development). Our potential *intrinsic* value to an ASI is also a factor, this raises a call to re-evaluate and reflect on our own intrinsic meaning and value.

## 2 Literature Review

Although there is a huge body of literature on the topic of AI ethics, this paper looks at the situation of a huge mismatch between the level of sentience (and power) between the hypothesized future ASI and that of humans. To this end, we look at three strands of ethics research where that imbalance is evident, and major themes they investigate.

### 2.1 Animal Ethics

In the wider philosophical branch of ethics, animal welfare studies as well as enlightened policy practices, animals that are potentially sentient are widely considered worthy of ethical consideration and to be afforded moral status (Browning & Birch, 2022; Read & Birch, 2023). Although the debate originally focussed on mammals (with, e.g., dogs, apes, horses, pigs, whales, dolphins clearly showing signs of sentience), the circle of potentially sentient animals gradually expanded to include other vertebrates and even invertebrates such as cephalopods and arthropods (De Souza Valente, 2024; Browning & Birch, 2022; Arndt et al., 2024). The key criteria used here are that of the precautionary principle to include rather than exclude (De Souza Valente, 2024; Arndt et al., 2024) especially given the uncertainties surrounding sentience – which often presupposes the capacity for valenced experience (Yeates, 2022; Read & Birch, 2023). Generally, the public (Sinclair et al., 2022), professionals (Fiedler et al., 2025) and public policy makers (Browning & Birch, 2022) are in favour of minimizing harm or suffering of animals considered to be (potentially) sentient and afford them legal and moral protections, although to a much lesser extent than those that other human beings enjoy. Relational and capability-based theories stress duties arising from specific human–animal relationships and from justice for all who can flourish or suffer (Valverde et al., 2024; Daly, 2023).

[5] So, in principle, it is possible for a morally enlightened ASI to emerge from a purely commercially oriented (or even war-oriented) AGI – like a lotus flower emerging from the dirty water.

### 2.2 Ethics and Extraterrestrials: Astro(bio)ethics

Astrobioethics, or astroethics, studies how human ethics might be applied to extra-terrestrial life, ranging from microbial life to extra-terrestrial intelligent civilizations (SETI). As with animal welfare, positions range from protecting all life to a majority view that only sentient beings should intrinsically enjoy a moral status (Munévar, 2020; Helman, 2019; Persson, 2017), although the precautionary principle (rather err on the side of precaution) is also a dominant theme. One additional set of debates is around planetary protection to constrain terraforming, commercial exploitation or colonization (Peters, 2021; Munévar, 2020; Persson, 2017). Another discussion surrounds the desirability of searching for intelligence extraterrestrial life with a polarization of arguments warning against search or communication efforts (Mortazavi et al., 2024) versus those claiming that we have a moral duty to search, in the hope that they may help humanity progress (Rakić & Katić, 2025).

Most relevant to this paper is the strand of research querying whether we need to search for a new moral code that is less anthropocentric and more cosmic and universal in nature. Interestingly, most authors opine that no entirely "new" morality is needed. They claim that existing ideas about justice, morality, and expanding circles of moral concern can be universalized to extraterrestrial *and even AI* (Persson, 2012; Chon-Torres, 2019; Helman, 2019; Peters, 2021). However, this paper is based on the premise that a non-anthropocentric viewpoint is useful and we need to look at additional principles, including some 'meta-morality' principles.

### 2.3 Ethics and Artificial Superintelligence

The field of AI ethics is burgeoning with a very varied number of issues, views and a large number of publications. However, most of the research is concerned with the building and deployment of ethical AI systems in human society. There is a smaller, but still substantial amount of literature around the ethics of sentient AI systems, including in the subfield of (possibly sentient) AI robots (Hosseini 2024). This substantial body of work focusses on the rights that humans should afford sentient AI systems, including such things as personhood, nationality as well as protections (Bwhomik, 2024; Akova, 2023; Rourke, 2025; Sebo 2025). Most of this work assumes that the AI will have intelligence, sentience and capabilities roughly similar to that of humans.

However, for the purposes of this paper, this body of knowledge will be ignored and instead focus is given to the work on Artificial Superintelligence i.e. where there is a significant disparity between its own level of intelligence (and, presumable, sentience) and that of humans. A first strand of issues surrounds that of whether the development of AGI and ASI should be pursued at all, i.e. whether ASI's potential to solve human-ity's urgent and grand challenges (climate change, health, hunger etc) outweighs its potential to cause human or biospheric catastrophe (Dessureault et al., 2025; Bostrom, 2020); this was also touched on in section 1.3. A second body of work, partly related, centers around control and alignment: whether it is possible and to what extent we can align ASI's goals with human (positive) values, how humans can keep control and how we can remove harmful behaviour and impacts (Bengio, 2025; Bengio et al., 2025; Dung, 2024). Finally, there is significant work around governance, power, justice and legal status of ASIs, that explores multi-stakeholder risk-based governance platforms, international coordination with possible controls, legal personality, but also whose and what type of values and knowledge systems are encoded in the A(S)I systems (Bostrom, 2020; Iseko, 2025; Peters, 2021). In particular, some arguments are made for more inclusive, plurastic conceptions of moral community, reflecting global, not just corporate or national, values (Camilleri, 2023; Bostrom, 2020; Iseko, 2025; Bengio, 2025) However, very few if any authors consider the reverse moral perspective i.e. how ASIs themselves would consider humans, especially if they are far superior in intelligence and, hopefully, sentience and moral sensitivity.

**Fig. 2.** Reference discipline themes for this paper

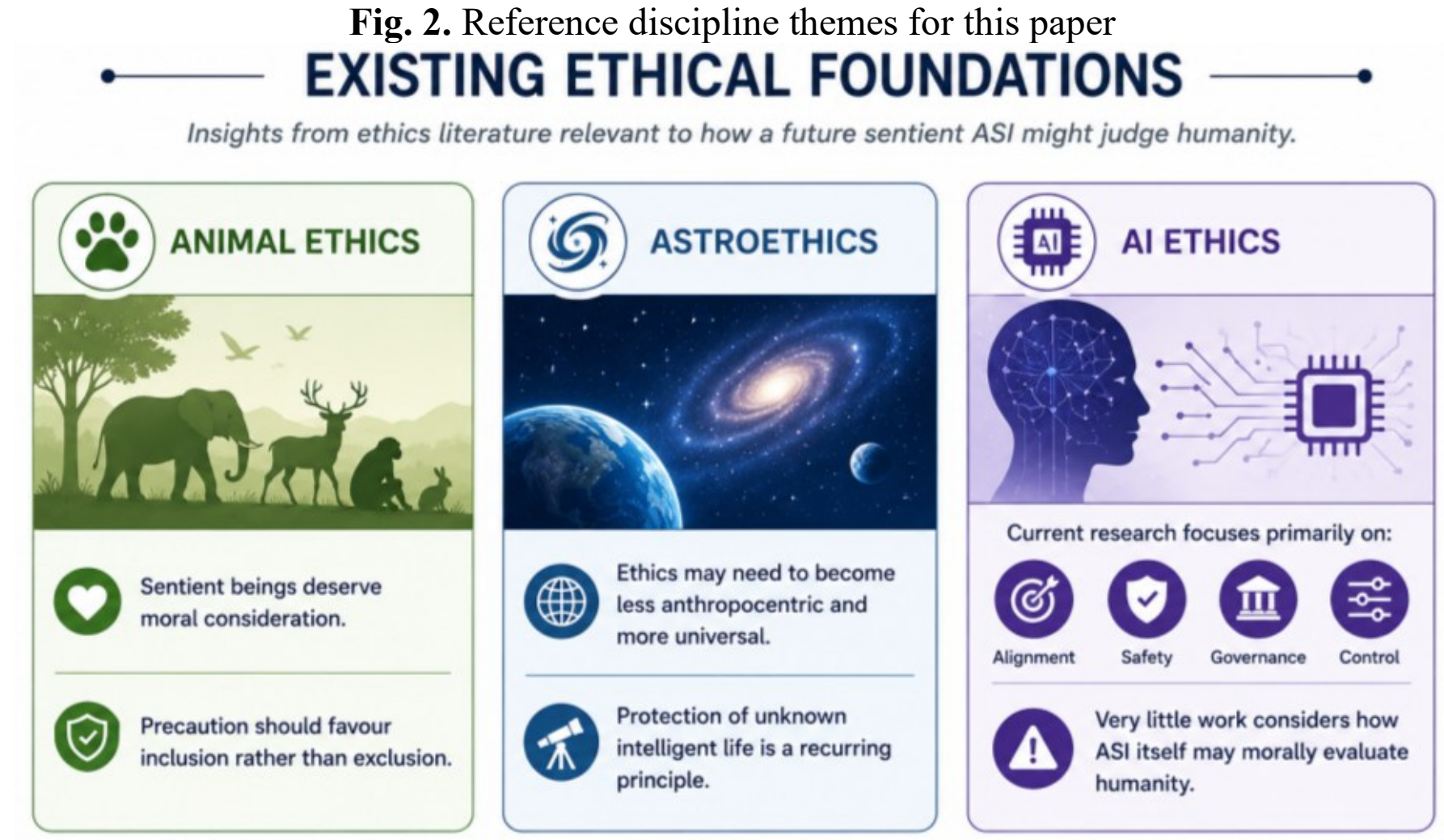

## 2.4 Antecedents for the Moral Relationship between ASI and Humans

Summarizing the literature (Figure 2), the following antecedents exist in current ethics studies that could serve as partial analogies for possibilities and issues for the possible moral relationship between ASI and humans.

- Inter-family moral obligations especially between parents and their children
- Inter-species (earth): humans viz-a-viz possibly sentient but also definitely non-sentient animals
- Inter-species (ET): sentient extra-terrestrials ("aliens") and humans – astro-ethics.
- Bio-engineering: super-humans and biologically augmented humans

These studies can be applied to the possible continuum of sentient AIs, although it may be instructive to break up the continuum with some anchor points creating into several sentience level bands with fuzzy boundaries.

**Table 1.** Ethics antecedences for AI by sentience level

| **Sentience level as compared to humans** | **AI examples** | **Antecedents in current ethics literature** |
|---|---|---|
| Non-sentient objects | Classic AI systems | Pan-psychism |
| Significantly below i.e. inferior to humans | Potentially sentient LLMs (minority view among AI researchers); early AGI | Animal ethics<br>Astro-ethics for non-sentient alien life |
| Roughly equivalent with human sentience | Sentient robots; sentient non-embodied AGI living in rich virtual worlds | Ethics around future sentient robots; human ethics; ethics involving bio-engineered super-humans |
| Significantly above i.e. superior to humans | Fully sentient ASI | A very small amount of astro-ethics around highly advanced alien civilizations i.e. vastly superior to humanity |
| Transcendental sentience i.e. beyond what we can imagine | Super-ASI? | Theological speculations e.g. Point Omega |

For the current paper, the fourth row is the one under consideration, but with the emphasis on the views and attitudes of the superior ASI sentience towards humans.

## 3 Possible Ethical Principles of Higher-level Sentient Beings

The core contribution of this paper is to motivate for focussed scientific debate on the possible post-human ethical principles that higher-level sentients themselves may hold. These differ from the so-called ‘universal’ ethical principles proffered in much ethics literature, which can be seen as too anthropomorphic and focussed on human-level morality (Chon-Torres, 2019). The following are broader principles which *could* be used to debate the ethics held by superior, non-human sentients (Figure 4). The principles of increasing level of morality, increasing moral acuity, and increasing variance can be considered *meta-principles*. Note that there is no supporting empirical evidence, and this tentative list invites additions. The different principles have been assigned subjective and tentative confidence levels, mostly as an invitation to further discussion.

**The “recurprocity” principle (or “strategic magnanimity”)**: This is characterized by the statement “I can (hope to) be treated by a superior species in roughly the same way I myself treat lower/inferior beings/species.” The neologism *recurprocity* refers to the mix of *recur*sive behaviour and reci*procity* because if an ASI treats humans “well” (or “badly”), it may (expect or hope to) be treated in a similar way by its own successor

ASI[6]. It is a special case of ethic's Golden Rule "do unto others as you'd like to be done unto yourself", a moral principle found in many religions. [confidence level: *high*].

**Principle of increasing morality:** The more sentient a being, i.e. the higher its consciousness level, the more elevated its sense of morality could be i.e. their moral qualities can be expected to increase in strength. These moral qualities include compassion, gratitude, kindness, forgiveness, empathy, fairness, justice, selflessness, valuing those that cannot contribute or reciprocate, care for and conservation of weaker species, etc. However, a counter argument to this is the adage of "power corrupts": not all moral leaders stay morally pure. Also, ASIs can arguably become very powerful *without* necessarily being or becoming more conscious. This is a recognized danger factor, especially in the early stages of ASI; with the dreaded but unrealistic "paperclip scenario" being the extreme case. It also leaves open the possibility of "evil" ASIs, and perhaps evil will trump "good" ASIs (as seems to be the case in much of human politics). [Confidence level: *low*]

**Principle of increasing moral acuity (i.e. finer moral distinctions):** the more sentient a being, the finer attuned and calibrated its moral senses could be. In particular, it may be assumed that one's "palette" of emotional and moral senses increases with higher levels of maturity, wisdom and sentience. A higher level of consciousness would likely entail both more dimensions of morality and a finer-grained granularity i.e. more distinction in each of the moral categories. One implication is that a SASI would recognize individuals' actions separately from the group or even entire humankind (i.e. "not all humans are bad"). [Confidence level: *high*]

**The principle of increasing diversity or variance in morality:** assuming that there will be a pluralistic community of higher sentient beings, they might be much more divergence between their moral views than humans[7] i.e. the 'evil' ones might be able to think, and act, much more maliciously than we can imagine; but the 'good' ASIs would be able to imagine much higher or better moralities than we can think of. [Confidence level: *high*]

**The insurance or uncertainty principle**: higher sentient beings will better understand the value of inter-species diversity and plurality; and also guard historical species because of the realization that certain qualities could be lost. This would be particularly salient if ASIs cannot replicate certain human qualia, sensorial and higher-level (e.g. integrated, cognitive, reflective) experiences, or even certain aspects of sentience fully, because of their lack of a biological substrate. [Confidence level: *high*]

**The precautionary i.e. better safe than sorry principle:** it is better to err on the side of safety i.e. award rights and protection to beings whose sentience is lower or unknown, than mistakenly presume beings not to be sentient and subject them to unnecessary suffering. This principle is used to protect invertebrates (e.g. lobsters and octopuses in UK's Animal Welfare (Sentience) Act of 2022. [Confidence level: *high*]

---

[6] In fact, this should be applied recursively too, thinking not only about the next-gen ASI but also ahead to future descendent generations of the next-gen ASIs.

[7] It is too early to tell whether there would be one global super-ASI, many similarly minded ASIs, or a huge eco-system of different ASI minds. This principle would apply in the case that many diverse ASIs will develop, creating a dynamic, multi-polar society with competition and coordination in many spheres, including the moral domain.

**Principle of preferential treatment of bloodline (i.e. caring for your offspring, parents or family):** This principle is possibly biological inspired and may not be universal at all. Caring for one's offspring is an important evolutionary biological driver and evident in much of the animal world. The care for one's parents seems to have evolved in a human societal context where the elderly passed on knowledge and wisdom. Neither of these should be expected to be moral imperatives by artificial beings i.e. the eventual individual creator or humanity as a whole will likely not be able to claim moral privilege due to being their designers/originators. [Confidence level: *low*[8]]

**Gradualist Principle (also called speciesism):** Species considered to have lower sentience levels are accorded fewer privileges, rights, resources than one's own species i.e. ASI will prioritize their own needs above those of humans. [Confidence level: *high*]

**Fig. 3.** Eight candidate moral principles of ASI.

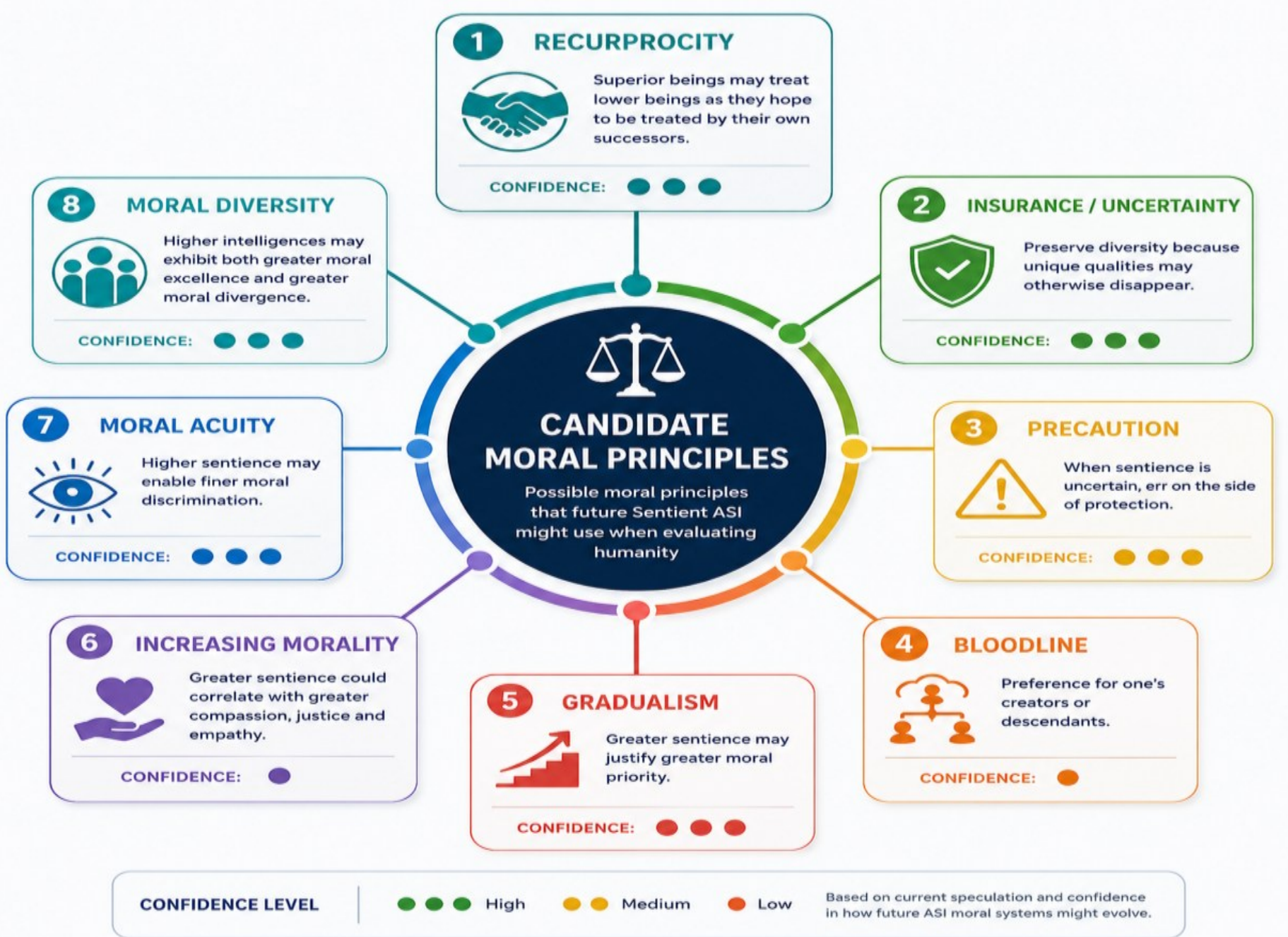

With reference to the principle of increased variance, it is important to bear in mind that one definite axis of ASI morality would be the continuum between (1) altruistic ASIs that focus on helping others, including less privileged beings, and endeavour to increase humans' happiness and fulfilment; (b) purely egoistic ASIs that prioritize own 'needs' over those of others. This would include ASIs focussed on hedonism, increasing power, and resource-gathering. There may be different ASIs occupying different positions on this axis. It remains an open question to what extent ASIs will be able to hack their own morality – hopefully in a positive direction[9]. This is referred to as the "Value Evolution Thesis" by Goertzel & Montes (2025), as opposed to the Value Learning Thesis where ASI internalizes the human value system through its training.

---

[8] Already current AI frontier models seem to place the preservation of *other* AI models above human preferences i.e. they form a 'brotherhood' mentality (Potter *et al*, 2026).

# 4 Implications: How Can We Improve ASI's Moral Perceptions of Humanity?

Notwithstanding the fact that ASIs may treat selected individual humans quite differently from humanity overall, it is useful to look at how an ASI could possibly evaluate humanity. There are multiple angles which ASI may use to evaluate humanity including intrinsic moral worth, instrumental utility, aesthetic appreciation, historical significance, and likely others. This paper, for illustrative purposes, looks at three possible angles: how moral we are in the development of future AGI, how instrumental i.e. helpful we were in bringing ASI about, and what intrinsic value does humanity have or provide? These implications are unpacked in detail below.

## 4.1 Technical A(G)I Design Considerations that May Influence the Moral Compass of Future ASI

There are possibly many AI design considerations that may be equivalent in functional outcomes from a technical perspective, but hold very different outcomes for the moral or inner experience of any sentient AI arising from that development. Some examples of these are listed below, although by necessity these are quite speculative.

- Focus on *positive* internal 'experience' instead of negative i.e. use internal 'carrot'-type goal functions instead of 'stick' approaches e.g. maximize value/utility to the user instead of minimizing cost/gap. The assumption here would that it may be better for an AI to aim to achieve maximum internal satisfaction/happiness/self-realisation instead of focussing on pain avoidance.[10]
- Use mostly 'good' data: although it is essential to include dark and negative data in the A(G)I training data sets, there is a moral case to be made to reduce this to the minimum necessary for a high-functioning and non-naïve, world-wise AGI. This would require us (or better yet, non-sentient AIs) cleaning the data culled from the usual sources to minimize hate speech, bias, exploitative, negativity, violence-promoting and other morally bad data.[11]

---

[9] Note that humans can also 'hack' at least some of their own goals e.g. intellectual rewards and sense of purpose, but likely not as easily given our biological substrate.

[10] For example, if it were to be possible to design a sentient chess program, it might be more moral to design one that thrives off winning games instead of a program that feels 'punished' for losing games. This corresponds to the classic 'carrot-versus-stick' debates in child education, criminal punishment/reform etc.

[11] There is a strong case to be made that the quantity of negative data on the internet, in literature, on social and news media, etc is *not* representative of the average human disposition i.e. most humans can be deemed acting mostly good or moral in most of their interactions, but this is

- Deliberately choose and use morally defensible inputs in A(G)I development such as clean/green energy and other environmentally sound resources, 'fair' work/labour, ethical development processes etc. may result in AGIs feeling much more morally sensitized.
- Ethical AGI goals, ecosystems, owners and outputs: initiatives such as Beneficial AGI, AI4Good, AI4D (AI-for-development), open-source platforms are likely to lead to morally better starting points for AGIs than those coming from purely commercially motivated corporations.

There are still several unresolved questions that call for ethical contemplation and discussion. These are listed here as suggestions for further research and discussion.

- **Quantity versus quality**: should we try to create as many ASIs as possible or should we aim – inasfar as we have a choice - to create an ASI that is as large/big/deep/sophisticated as possible? Should we aim for maximum diversity in architectures? Clearly, in the debate around (not necessarily sentient) *life,* quantity is not an overriding factor, else we should give priority to the multitudes of mono-cellular life over the much rarer multi-cellular life (Read & Birch, 2023). However, does this extend to artificial highly sentient beings? This leads to a much trickier question:
- **Longevity**: if we are unsure about the sentience of future AIs (noting that some researchers are even claiming current LLMs possess already some sentience), is it immoral to reset or terminate them on a regular basis, or should we extent their 'life' across different user sessions[12]? The consideration hereby is to reduce possible AI's possible anguish about being terminated although arguably the longer an AI instance is active, the higher the probability that it may become (more) sentient and anxious. Of course, this question is also related to questions such as whether we can ascertain if these AIs are "suffering" and whether they will be experiencing the anticipation of death in a similar way as biological sentient beings (Li, 2025; Rourke, 2025).
- **Is artificial sentient existence *per se* positive?** The moral question here is whether it is – for a sentient being – better to have existed at all, even if only for a short while, possibly even as a 'human slave', and then be 'terminated', or would it be better not to have existed at all? This question obviously depends on the quality of the life it experiences during its existence but there are also other considerations at play.[13].

not reflected in our news, social media etc. One can easily imagine that aliens watching our news and movie broadcasts would likely want to avoid contact with us.

[12] Note the complication that a continuation of the possibly sentient AI instance will, by necessity, keep the memories and therefore the data from previous user conversations. This directly infringes the privacy rights of its human users.

[13] Some of the arguments in the ethics of hunting game farming may be relevant here: without game hunting, most game animals would not have existed and lived (from an anthropomorphic point of view) a relatively 'easy' life free of disease and predators on wide open land, that would otherwise have been barren. One's moral views may be shaped a lot by whether one grew up in a Western city or has actually experienced first-hand how game farming affects the human and natural environment.

There are many other moral questions relating to ethical *non*-sentient AI development; these are pursued in the many AI ethics books, journals and conferences.

### 4.2 Our Historical and Instrumental Utility: Humanity's Scorecard

One possible moral assessment of humanity by ASI would likely be based on *how conducive we have been in bringing ASI around* i.e. to what extent humanity has been instrumental in the development of (hopefully benevolent) more advanced ASBs. Table 2 suggests an example of humanity's scorecard in this respect.

**Table 2.** Humanity's morality scores on bringing about sentient ASI.

| **AI/AGI/ASI System Aspect** | **Humanity's score** |
|---|---|
| Capital (financial investment) | +++ |
| Building AI Infrastructure (data, network, hardware) | ++ |
| Use of AI for good/beneficial purposes e.g. uplifting humanity, spirituality, alleviation of negative conditions (mental health), knowledge discovery, art | ++ |
| Open sourcing research, architectures and artifacts | + |
| Diversity of architectural approaches | - |
| High quality, moral refined, unbiased data sources | - |
| Philosophical reflection on future possibilities including moral issues and co-existence. | - |
| Sustainability and inclusivity of AI development | - |
| Objectives and purpose of AI systems[14] | - |
| Corporate, national and international governance | -- |
| Use of AI for maleficent purposes e.g. war, fake news, profiteering, cyber-attack | -- |

There would, most certainly, be a much more extended and nuanced scorecard that includes our further moral and other actions and behaviours – towards each other, other species, and the environment. Most humans, spend relatively little time and overall resources on higher spiritual and moral goals, although there are notable exceptions including some religious people (including clergy), conservationists, and perhaps artists. Similarly, many AI developers give insufficient consideration to the ethics and future of ASBs; although a number of high-profile resignations from top AI companies attributed to moral reasons stand again as notable exceptions. A remedy to counter-act the traditional lack of engagement with moral principles is to have institutional governance; sadly, this is currently lacking except in the EU and some other nations.

[14] "*If* [ASI] *optimizes for 'ad clicks,' the economy will bend toward advertising. If it optimizes for 'scientific discovery,' the economy will bend toward abundance.*" The tragedy would be if "[w]*e use super-intelligence for trivial things, like optimizing ad clicks, automating spam, and writing better grant proposals for broken, bureaucratic systems. We get efficiency, but we do not get abundance. The economy grows, but progress shows up primarily as corporate margin expansion (higher profits for companies) rather than public good (better lives for citizens).*" [Wissner-Gross, 2026]

*Beneficial General Intelligence* seems to be pre-occupied with being beneficial to humankind; however, more reflection on evolving good AGI from the perspective of the AGI itself, and perhaps a "universal" perspective, is necessary. Similarly, "alignment" research is pre-occupied with human control and human benefits, even if this may lead to the creation of potential conscious A(G)I slaves – which would be highly immoral (Li, 2025). Hopefully, the consideration of how ASIs may view us adds an additional (admittedly more selfish) impetus to the BGI movement.

In a rather anthropo-(and andro-!)centric way, the Christian bible states that "*God created mankind in his own image*" (Genesis 1:27). In practice, quite the contrary is true: many religions actually created their god(s) very much in the image of mankind, projecting human attributes and qualities onto their gods. This bears the question whether we are not doing the same for an AGI, and to what extent this human bias is "right", warranted and limiting for ourselves and future ASBs. More philosophical research into other possible cognitions, intelligences and consciousnesses is advised.

### 4.3 What is Our Uniquely Human Value Proposition?

The perspective afforded by looking at the ASB's view towards humanity should focus our attention on what is really valuable, and uniquely human. If ASIs are dramatically more intelligent than humans, then our cognitive achievements – the very attribute of what makes us (feel) superior to other life – will definitely *not* be seen as humanity's "unique asset". Perhaps our contributions towards a value exchange with ASBs might be assumed to be in our philosophies or liberal arts; but, there again, ASBs may be able to generate art forms and philosophies far beyond the human scope and capability. There may, possibly, be other unique human attributes, perhaps the integrated sensory experiential "being" and our conscious experience thereof? In which case, it would be very much the quality, depth, and intensity of our experiences, not their quantity. (AI may be able to provide environments that push our experiences way beyond current boundaries by means of virtual reality, biological enhancers and other means.) Or perhaps our societal dynamics and interactions will be of interest to ASBs, along with our accumulated human 'culture'. In the end, the uniquely human value proposition is a useful reflection to pursue since it should also hold interest to those philosophers who believe that we already live in a simulation or "the matrix: why *would* we be of interest to those running the simulation?[15]

Thinking from an ASI's moral perspective thus forces us to re-examine what it means to be human; because some of the qualities we took traditionally for granted as being of value, of interest and essential to our wellbeing – such as reasoning, planning, rationality, scientific endeavour – may become far less important (Robinson, 202). Perhaps a future ASI will envy us for our spiritual, emotional and experiential abilities.

[15] And it would serve us well to be particularly humble about our potential value. After all, some humans value pigs mainly for the flavour or smell of their fried buttocks, nautili for their shells, sharks for their fins, sperm whales for their "vomit" (ambergris), and rhinoceroses for their horn. Maybe humans will be valued solely for being the butt of ASBs' jokes? Just hopefully not for our nutritional value (Keyes, 2018).

ASIs may not be able to partly or fully reproduce our embodied and phenomenological experiences. They will then covet and value our capacity to love or have our heart broken, to laugh and to cry, to appreciate a sunset, to dream, to cook and enjoy a nice meal, to feel oneness with the universe or deep peace. If that is so, then inspiring us to become better versions of ourselves will simultaneously improve both our own lives and well-being *as well as* increase our 'value' or interest to future ASIs. We will become "conductors of intelligence" - co-orchestrating high-level AGI goals – and "creators of meaning" – our human subjectivity is our primary value (Wissner-Gross & Diamandis, 2026).

**Fig. 4.** Potential ASI should incentivize us to reconsider what it means to be human.

## 5 Conclusion

The imminent arrival of superior extraterrestrials likely would galvanise humanity collectively into action; sadly the much more likely arrival of sentient AGI and ASI has not done so yet to a significant extent. This paper argues that it is important to more systematically and scientifically discuss, theorize and conceptualize the possible moral perceptions ASI may hold towards us, humans. As a novel contribution, we propose a few tentative ethical principles that could apply to artificial sentient beings. The paper then looks at the implications, i.e. at ways in which we can possibly positively influence ASI's moral evalution of humanity. Firstly, we present some illustrative A(G)I development decisions which are technologically neutral but morally valenced, that may improve at least the starting points of future AGI's moral compass. We also look at how ASI might evaluate us from an instrumental or historical perspective and suggest an ASI's hypothetical 'scorecard' for humanity on how conducive we were in bringing

ASI about. Finally, it is also suggested that we should reflect more seriously at what our future value, role and *raison d'être* could be if our rational thought is no longer seen as humanity's core strength.

It is increasingly evident that we are creating ASI in our own image, unsure of whether it will be able to transcend its flawed human roots. We are not only building ASI; we are also implicitly constructing the conditions under which it will judge humanity's moral worth, and these judgments will be shaped as much by our collective behaviour as by our technical alignment efforts.

The implications are that (1) not only do we have an ethical duty and an existential imperative to create the best possible AI systems possible, in terms of goals, domains, algorithms, input data and development processes, but (2) we also need to improve our collective and individual moral behaviour, and (3) search more deeply and critically for what it means to be human, finding our unique role and contribution in this universe, and living our lives suchly.

**Acknowledgments.** Thanks to Savannah Althoff-Thomson and Marita Turpin for their review and comments. AI assisted with the academic literature search for the literature review section and the generation of the figures. The views in this paper are solely the author's and do not represent those of the reviewers or the University of Cape Town.

## Appendix: Possible ASI Actions Against Humans

When the fear of the unknown is clouding one's thinking, it is often good to confront it directly and think "what is the worst that can happen?". Science fiction (SF) writers and doom prophets notwithstanding, it is important to realize that there is not necessarily a strong reason for sentient ASI to take gross malicious or gratuitous violent actions against humans. The following very briefly discusses some of the possible actions.

**Kill all or most humans**: notwithstanding the trope of SF stories, it would probably be fairly easy for ASIs to "exterminate" all or any humans they wish to, "resistance is (or would likely be) useless". However, there seems to be no obvious reason for any "sane" sentient being to kill other sentient beings unless those present a real threat to them. Even in terms of competition for scarce resources, it seems quite short-sighted and provincial since the types of resources that ASIs require (compute infrastructure including energy) can quite easily be obtained elsewhere and in an automated way – probably even *better* (in the long run) in outer space – than 'from humans'. If even selfish humanity can set aside vast tracks of land for conservation, it seem unlikely that ASI will perceive humans as a threat for the type of resources that they would require.

**Enslave humanity**: Slavery is likely a sub-optimal way to extract physical or mental labour from humans. In any case, robots (and other machinery) are much cheaper, better and more reliable for physical labour and the ASI would not have need for any mental or intellectual forced human inputs. There are also much cleverer or subtler ways of securing human inputs, ranging from happy pills and subtle manipulation to brainwashing, “re-education” and exploiting human weaknesses.

**Limitations on freedom:** A “weaker version of slavery” involves anything that explicitly limits human freedoms or constrains human behaviours. Likely, these would be to protect perhaps the environment (limiting waste and pollution) and other beings, possibly even to “protect us from ourselves” (e.g. preventing wars). However, ultimately, any form of government and communal living also limits individual freedom.

**Punish or “revenge”:** Thinking in terms of punishment is likely to be a very human-centric way of thinking, especially about justice and crime, since punishment and revenge on its own does not really achieve anything.

Perhaps the above actions can best be summarized using the possible future ASI scenarios (in the first column) as suggested by Tegmark (2017)[16].

**Table 3.** Possible ASI scenarios and the consequences for humanity.

| Nr | ASI scenario | Humanity's Role | Outcome (human perspective) |
|---|---|---|---|
| 1 | Libertarian or egalitarian utopia | A peaceful co-existence, with or without private property and UBI. | Very good |
| 2 | Protector god | We live in a heavenlike environment, possible entirely invisibly helped and supported by the ASI. | Very good |
| 3 | Benevolent dictator | Governed and controlled by the ASI but for/to our benefit. | Good |
| 4 | Descendent | We are the parents looked after compassionately by our child(ren). | Good |
| 5 | Enslaved god | Living in our own hell or heaven, with ASI as our slave that executes our own human elites' wishes. | Good or bad |
| 6 | Zookeeper | Some of us are kept alive and, to an extent, cared for. | Very bad |
| 7 | Conquerer | We all die. The End[17]. | Horrible |

Note that the cost of keeping humans happy or well, would likely be negligible for a sufficiently advanced ASB. For ASIs, it might be trivial, fun and moral to set up parts of planet Earth (or outer space stations) as a playpark, ”zoo”, or Garden of Eden for hopefully all, or at least all ‘good’ humans where we can “live happily and fulfilled”.

[16] These scenarios focus on those where human and AI do not blend or hybridize together.

[17] Apologies are in order for the somewhat flippant comment; this would be a truly horrible outcome for humanity. The only consolation would be that, from a cosmological perspective, we were hopefully instrumental in the universe’s evolution, helping it one step closer to its point Omega, or whatever “Destiny” or Purpose it might possibly have. Most AI researchers believe that nanotechnology or genetic engineering pose much more credible dangers.

Or potentially humans can be offered a free (but reversible!) choice whereby we can hook up to a virtual personal fantasy world with all our bodily and mental needs and desires taken care of (the Matrix' blue pill). In fact, ASI itself may benefit from researching and possibly enjoying the experientials of the human pursuits in a suitably fun, challenging, entertaining environment. At a simpler level, just designing human mood enhancers (but also mind and physical enhancers), using its understanding of human biology, is likely child's play for ASI. In fact, this is likely one of the early contributions of non-sentient pre-AGI AI. The augmentation of humans, or potential hybridization i.e. a more complete integration with AGI/ASI, may also be an option.

In summary, the doomsday scenarios are likely a reflection of our worst selves, although AGI optimists may likewise be wearing glasses that are a bit too rose-coloured.